\documentclass[letterpaper, 10pt, conference]{ieeeconf}  
\IEEEoverridecommandlockouts
\usepackage{color}
\usepackage{amsmath}
\usepackage{mathptmx} 
\usepackage{times} 
\usepackage{amsmath} 
\usepackage{amssymb}  
\usepackage[caption=false,font=footnotesize]{subfig}
\usepackage{calrsfs}
\DeclareMathAlphabet{\pazocal}{OMS}{zplm}{m}{n}
\usepackage{graphicx}
\usepackage[off]{auto-pst-pdf}
\usepackage{psfrag}
\usepackage{breqn}

\title{\LARGE \bf
  Non-Central Catadioptric Cameras Pose Estimation using 3D Lines*
}

\author{Andr\'{e} Mateus, Pedro Miraldo and Pedro U. Lima
\thanks{All authors are with the Institute for Systems and Robotics, Instituto Superior T\'{e}cnico, Universidade de Lisboa, Torre Norte - 7 Piso Av.Rovisco Pais, 1 1049-001 Lisboa, Portugal.\newline {E-Mail:~\tt\small~pmiraldo@isr.tecnico.ulisboa.pt}}%
}

\begin{document}

\maketitle
\thispagestyle{empty}
\pagestyle{empty}

\begin{abstract}
  In this article we purpose a novel method for planar pose estimation of mobile robots. This method is based on an analytic solution (which we derived) for the projection of 3D straight lines, onto the mirror of Non-Central Catadioptric Cameras (NCCS). The resulting solution is rewritten as a function of the rotation and translation parameters, which is then used as an error function for a set of mirror points. Those should be the result of the projection of a set of points incident with the respective 3D lines. The camera's pose is given by minimizing the error function, with the associated constraints. The method is validated by experiments both with synthetic and real data. The latter was collected from a mobile robot equipped with a NCCS. 

\end{abstract}

\section{Introduction}

The ability of a robot to estimate its absolute pose and/or localize itself in the environment is a fundamental task for an autonomous robot. The pose estimation problem consists in finding the rigid transformation, between the robot's frame and the world coordinates system, which is defined by a rotation and a translation. In this work, a robot is equipped with a NCCS on-board, which is used to estimate its pose. Catadioptric devices have been in used in some application is robotics, an example is robot competions, \cite{marques:2000}.

The majority of vision-based pose estimation methods proposed in the literature focus on perspective cameras, \cite{book:hartley:2004}. Examples of methods are: \cite{article:araujo:1998}, \cite{article:moreno:2007} using non-minimal number of known 3D points; \cite{article:ansar:2003} non-minimal solutions using 3D lines; and \cite{article:ramalingam:2011}, \cite{article:haralick:1994} for minimal solutions using both points and lines. The widespread use of this type of cameras is due to their simplicity and well-known mathematical model. However, their field of view (FOV) is limited. In order to overcome that limitation, the focus is increasingly shifting towards other imaging devices, which ensure a wider FOV, the most notable are the catadioptric cameras, \cite{article:nayar:1997}. These cameras combine quadric mirrors with perspective cameras for increased FOV. Some of these devices were built to comply with the central projection model, e.g. \cite{article:baker:1999}, \cite{article:geyer:2000}. However, in general (and in practice) this constraint is not verified. Thus, the catadioptric camera systems are, most of the times, non-central cameras, i.e., they do no verify the central projection model \cite{article:swaminathan:2006}.

The problem of absolute pose estimation, based on general non-central camera models, as been addressed by Chen and Chang, \cite{article:chen:2002}, Schweighofer and Pinz, \cite{article:pinz:2008}, Nister and Stewenius, \cite{article:nister:2004} and Miraldo and Araujo, \cite{article:miraldo2:2014} for the known matching between 3D points and their correspondent image pixels. This problem was also addressed for known 3D lines at Miraldo et al.,~\cite{article:miraldo:2015}, for the 3D pose, and at Miraldo and Araujo, \cite{article:miraldo:2014}, for the planar pose (the problem addressed in this work). 

The problem of the projection of 3D points onto mirrors and, consequently, to images of non-central catadioptric cameras as been studied by some authors, in the last few years. For instance at \cite{article:goncalves:2010}, Gon\c{c}alves proposed an iterative solution to this problem. Later, Agrawal et. al. \cite{article:agrawal:2011}, proposed an exact projection model (but still iterative for general configurations) for NCCS. They derived a forward projection equation, with no restrictions in the camera's location, where the projection point on a rotationally symmetric quadric mirror can be found (in general) by solving an $8^{th}$ degree polynomial equation. However, in practice, it is useful to use other features, such as 3D straight lines. Since the lines are an one dimension object, the association between their features in the world and its respective images is easier and, thus, can be used for a wide range of applications.

This work is two-sided, we first derive the equation, which represents the projection of a 3D straight line onto the mirror's surface of non-central catadioptric cameras (henceforward denoted as reflection curve). We concluded that the curve can be analytically represented by a $10^{th}$ degree polynomial equation. Then, we address the planar pose estimation problem by means of an objective cost function. This was obtained by rewriting the reflection curve as a function of the rigid transformation parameters. The objective cost function is then applied to a set of mirror points (in the camera reference frame), which belong to the reflection curve of known 3D lines (in the world frame). The solution is found by minimizing the sum of the absolute value of the cost function for each point of each line. The methods are validated through synthetic data in different types of mirrors. The pose estimation method is also validated with real data, from a NCCS mounted on top of a mobile robot.

Throughout this article, we denote vectors by lowercase bold letters, e.g. $\mathbf{b}$, matrices are denoted by uppercase bold letters, e.g. $\mathbf{A}$, and regular lowercase letters represent zero dimension elements. The symbol $\sim$ is used to represent up to a scale factor equations. The superscripts $(\pazocal{W})$ and $(\pazocal{C})$ represent elements in the world and camera frames respectively. Capital greek symbols represent 3D elements, e.g. $\Gamma(x,y,z)$, lowercase symbols represent 2D elements, e.g. $\gamma(y,z)$, with the exception of the symbols $\lambda$ and $\theta$, which represent problem variables.

This paper is structured as follows: Section~\ref{sec:line_proj} presents the reflection curve derivation; in Section~\ref{sec:pose} we describe the pose estimation method. Sections~\ref{sec:syn_results} and \ref{sec:real_results} present the results using synthetic and real data respectively. Finally conclusions are presented in Section~\ref{sec:conclusions}

\section{3D Line Projection onto Non-Central Catadioptric Cameras} 
\label{sec:line_proj}

\begin{figure}
  \centering
  \psfrag{Vi}[c][c]{\small$\mathbf{v}_{i}(\lambda)$}
  \psfrag{Vr}[c][c]{\small$\mathbf{v}_{r}(\lambda)$}
  \psfrag{M}[c][c]{\small$\mathbf{m}(\lambda)$}
  \psfrag{Line}[c][c]{\small$\mathbf{p} \doteq \mathbf{q} +\lambda \mathbf{d}$}
  \psfrag{Gamma}[c][c]{\small$\Gamma$}
  \includegraphics[width=0.9\columnwidth]{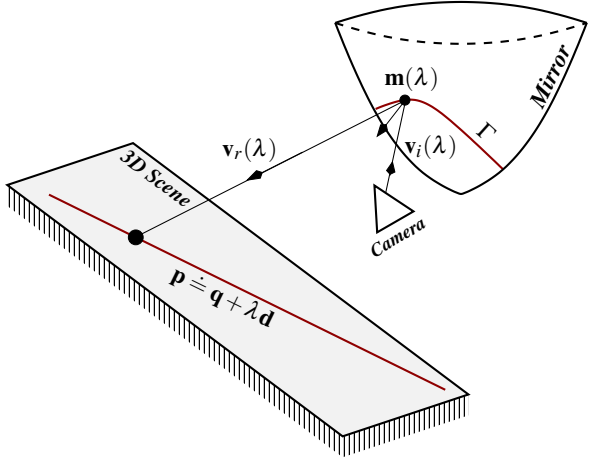}
  \caption{Representation of a 3D lines and its respective reflection curve on the mirror (in red).}
  \label{fig:proj}
\end{figure}

In this section we derive the equation that represents the reflection curve of a 3D straight line on the mirror. The solution should be of the type $\Gamma(x,y,z)=0$, in order to verify if the point $(x,y,z)$ belongs to the curve, Fig.~\ref{fig:proj}. This derivation was based on the same constraints used at Agrawal et al. \cite{article:agrawal:2011}

Consider a catadioptric system, which consists of a perspective camera, centered at $\mathbf{o} = (0, o_y, o_z)\in \pazocal{P}^3$, and a rotational symmetric quadric mirror. Without loss of generality, let us consider the $z$--axis as the mirrors rotation axis. Thus the mirror can be described by
\begin{equation}
  \Omega(x,y,z) \doteq x^2 + y^2 + Az^2 + Bz - C = 0,
  \label{eq:mirror}
\end{equation}
where $A$, $B$, and $C$ are the quadric mirror parameters. Consider also a 3D straight line defined by a point, $\mathbf{q}$, and a direction, $\mathbf{d}$, so that any point on the line can be given by
\begin{equation}
  \mathbf{l} \doteq \mathbf{p}(\lambda) = \mathbf{q} + \lambda\mathbf{d}, \text{ for some } \lambda.
  \label{eq:line}
\end{equation}
From the Snell's law, we get two well-known constraints:
\begin{itemize}
  \item{Any point in a 3D line, its reflection point on the mirror, and the camera's effective view point define the reflection plane, $\pi$;}
  \item{The angle between the incoming rays and the normal at the mirror's surface is equal to the angle between reflected rays and the normal.}
\end{itemize}      
The former can be written as
\begin{equation}
  \pi \doteq \mathbf{p}(\lambda) \cup \mathbf{m}(\lambda) \cup \mathbf{o},
  \label{eq:rplane}
\end{equation}
where $\mathbf{m}(\lambda) = (x(\lambda),y(\lambda),z(\lambda))$, represent a reflection point, for some $\lambda$.

The normal vector can be computed by taking the gradient of \eqref{eq:mirror}, resulting in
\begin{equation}
  \mathbf{n} = \nabla \Omega(x,y,z) = [x\ y\  Az + B/2]^T.
  \label{eq:normal}
\end{equation}
Since the normal vector at the reflection point on the mirror lies on the reflection plane, any point defined by $\mathbf{k} = \mathbf{m}(\lambda + \nu\mathbf{n}$, also belongs to the plane. Setting $\nu = 1$, the point
\begin{equation}
  \mathbf{k} = [0\ 0\  z-Az + B/2]^T,
\end{equation}
can be defined. Since this point lies on the reflection plane, it can be defined by $\pi \doteq \mathbf{p}(\lambda) \cup \mathbf{k} \cup \mathbf{o}$. Computing the plane equation from the previous definition and solving for $x$, we obtain
\begin{equation}
  x = -\frac{c_3^2[y,z]\lambda + c_4^2[y,z]}{c_1^1[z]\lambda + c_2^1[z]},
  \label{eq:solve_for_x}
\end{equation}
where $c_i^j[.]$ is a $j^{th}$ order polynomial equation. Replacing \eqref{eq:solve_for_x} in the mirror equation \eqref{eq:mirror}, and rearranging it, we get 
\begin{equation}
  c_5^4[y,z]\lambda^2 + c_6^4[y,z]\lambda + c_7^4[y,z] = 0.
  \label{eq:reflection_plane}
\end{equation}

In order to get an analytical equation for the projection of lines, the parameter $\lambda$, must be removed. To achieve that, we take the advantage of the fact that the reflected ray, $\mathbf{v}_r(\lambda)$, must go through the respective 3D line point, $\mathbf{p}(\lambda)$, such that
\begin{equation}
  \mathbf{v_r}(\lambda) \times (\mathbf{p}(\lambda) - \mathbf{m}(\lambda)) = 0.
  \label{eq:reflected}
\end{equation}
Besides, from Snell's law, one can derive 
\begin{equation}
  \mathbf{v_r}(\lambda) \sim \mathbf{v_i}(\lambda) - 2\mathbf{n}\frac{\mathbf{v_i}(\lambda)^T\mathbf{n}}{\mathbf{n}^T\mathbf{n}}
\end{equation}
and, since the scale of $\mathbf{v_r}(\lambda)$ is not important, it can be rewritten as
\begin{equation}
  \mathbf{v_r}(\lambda) \sim 4(\mathbf{n}^T\mathbf{n})\mathbf{v_i}(\lambda) - 8\mathbf{n}(\mathbf{v_i}(\lambda)^T\mathbf{n}).
  \label{eq:vr}
\end{equation}
Finally the incident ray, $\mathbf{v_i}(\lambda)$, can be written as
\begin{equation}
  \mathbf{v_i}(\lambda) \sim \mathbf{m}(\lambda)-\mathbf{o}
  \label{eq:vi}
\end{equation}

Replacing \eqref{eq:vi} and \eqref{eq:normal} in \eqref{eq:vr}, and then this last one in \eqref{eq:reflected}, three linear dependent equations are obtained. Thus, they represent one single constraint. For simplicity sake, we take the equation independent of the variable $x$. Solving the chosen equation for $\lambda$, we get
\begin{equation}
  \lambda = -\frac{c_9^3[y,z]}{c_8^3[y,z]}.
  \label{eq:lambda}
\end{equation}
Replacing $\lambda$ on \eqref{eq:reflection_plane}, after some simplification, we get
\begin{multline}
  \gamma(y,z) = c_5^4[y,z](c_9^3[y,z])^2 - \\ c_6^4[y,z]c_8^3[y,z]c_9^3[y,z] + c_7^4[y,z](c_8^3[y,z])^2 = 0,
  \label{eq:gama}
\end{multline}
where $\gamma(y,z)$ is a $10^{th}$ order polynomial. Moreover replacing \eqref{eq:lambda} in \eqref{eq:solve_for_x} we obtain
\begin{equation}
  x = -\frac{-c_3^2[y,z]c_9^3[y,z] + c_4^2[y,z]c_8^3[y,z]}{-c_1^1[z]c_9^3[y,z] + c_2^1(z)c_8^3[y,z]} = \frac{c_{11}^5[y,z]}{c_{10}^4[y,z]}.
  \label{eq:x_solved}
\end{equation}
In conclusion, a point $(x,y,z)$ in the mirror belongs to the reflection curve of the 3D straight line (defined in \eqref{eq:line}), if and only if \eqref{eq:gama} are verified.

\section{Planar Pose Estimation from 3D Straight Lines}
\label{sec:pose}

\begin{figure}
  \centering
  \includegraphics[width=1\columnwidth]{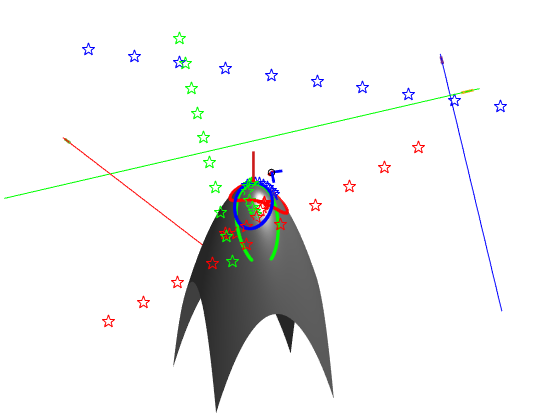}
  \caption{Representation of the proposed problem with $N=3$ and $M=10$. Notice that the 3D lines and the respective 3D points (starts) are not incident.}
  \label{fig:prob_form}
\end{figure}

In the previous section, an analytical solution for the projection of 3D straight lines onto the mirror of NCCS was derived. In this section that equation is rewritten, in order to obtain an error function for the estimation of the absolute planar pose.

The problem of pose estimation consists in finding the rotation matrix $\mathbf{R}\in\pazocal{SO}\left(3\right)$ and translation vector $\mathbf{t}\in\mathbb{R}^3$, which define the rigid transformation between the world and camera's reference frames. Keep in mind that only the planar pose estimation is considered, we have three degrees of freedom. Assuming that the robot/camera is moving on a plane parallel to the $xy$--plane, those degrees of freedom correspond to a rotation angle ($\theta$) around one axis (the $z$--axis) and the other two to the translation ($t_x$ and $t_y$). In this scenario, the rotation matrix and translation vector are
\begin{equation}
  \mathbf{R} = \begin{bmatrix}
    c \theta & -s \theta & 0 \\
    s \theta & c \theta & 0 \\
    0 & 0 & 1
  \end{bmatrix}
  \text{and }
  \mathbf{t} = \begin{bmatrix}
    t_x \\
    t_y \\ 
    c^{te}
  \end{bmatrix},
  \label{eq:rt_para}
\end{equation}
where $c\theta$ and $s\theta$ represent $cos(\theta)$, and $sin(\theta)$ respectively and $c^{te}$ is a known constant. Thus the unknowns of the problem are the rotation angle $\theta$, $t_x$ and $t_y$.

Let us consider a set of $N$ known straight lines in the world frame, $\mathbf{l}_i^{(\pazocal{W})}$ (filled lines in open space on Fig.~\ref{fig:prob_form}), for $i=1,\dots,N$, which are not aligned with the camera's coordinate system. Consider also a set of $M_i$ pixels $\mathbf{u}_{i,j}$, for $j=1,\dots,M_i$, which correspond to the $j^{\text{th}}$ point in the image of the $i^{\text{th}}$ straight line. Since the NCCS is considered to be calibrated (we known the projection matrix and the parameters of \eqref{eq:mirror}), the reflection points on the mirror $\mathbf{m}_{i,j}^{(\pazocal{C})}$ (star points on the mirror, Fig.~\ref{fig:prob_form}), correspondent to the pixels $\mathbf{u}_{i,j}$, are easily obtained. Notice that these reflection points are represented in the camera's coordinate system, while the lines are represented in the world's coordinate system. To compute these points, one needs to re-project the pixels and intersect the respective camera's projection line with the mirror. Given that the reflection curve equation (derived in Section~\ref{sec:line_proj}) assumes that both the lines and the mirror points are in the same reference frame, we cannot apply \eqref{eq:gama} directly.

In order to have both the lines and the mirror points on the same reference frame, there are two options. The first consists of having the lines position fixed and apply a rigid transformation to the camera system (both the mirror and the perspective camera). However, the respective formulation for the problem, would not be trivial. The second option consists in applying a rigid transformation to the lines, having the camera coordinate system fixed. Considering both lines and mirror points on the camera reference frame, the derivation is simpler.

Let us consider the second option (as shown in Fig.~\ref{fig:prob_form}). Applying the rigid transformation to the lines defined by \eqref{eq:line}(filled lines in open space on Fig.~\ref{fig:prob_form}), one gets
\begin{equation}
  \mathbf{p}(\lambda)^{(\pazocal{C})} = \mathbf{R}\mathbf{p}(\lambda)^{(\pazocal{W})} + \mathbf{t} = \lambda \mathbf{R}\mathbf{d}^{(\pazocal{W})} + \mathbf{R}\mathbf{q}^{(\pazocal{W})} + \mathbf{t},
  \label{eq:tline}
\end{equation}
where $\mathbf{p}(\lambda)^{(\pazocal{C})}$ represents a line in the camera frame (in Fig.~\ref{fig:prob_form} are represented by the star points in 3D). Now, that both mirror points and 3D lines are in the same coordinate system, \eqref{eq:gama} can be rewritten. The goal of this reformulation is to estimate the transformation applied to the lines in the world coordinate system, in order to their reflection curves intersect the mirror points in the camera coordinate system. By replacing \eqref{eq:tline} in \eqref{eq:rplane} and following the steps of the derivation described in Section~\ref{sec:line_proj}, we get a function
\begin{equation}
  \gamma_r(y,z,\theta,t_x,t_y) = 0,
  \label{eq:gamar}
\end{equation}
which is a function of not only the mirror point coordinates, but also of the rigid transformation parameters.

Given that we know a set of mirror points of the transformed lines, we have a set of $y$ and $z$ parameters, which means that we can consider that \eqref{eq:gamar} depends only on the rigid transformation parameters (becoming $\gamma_r(\theta,t_x,t_y)$). To simplify the rotation parameter (which include non-linear sine and cosine functions), we consider as unknowns the variables $c\theta$ and $s\theta$. Since these parameters are not independent, we have to take into account the following constraint 
\begin{equation}
  g_1(c\theta,s\theta) = c\theta^2 + s\theta^2 = 1.
  \label{eq:constraint}
\end{equation}
As a result, the final equation for the reflection curve, as a function of the rigid transformation parameters, is given by
\begin{dmath}
  \gamma_r(c\theta,s\theta,t_x,t_y) = c_{12}^4[c\theta,s\theta,t_x,t_y] + c_{13}^3[c\theta,s\theta,t_x,t_y] + c_{14}^2[c\theta,s\theta,t_x,t_y] + c_{15}^1[c\theta,s\theta,t_x,t_y] + c_{16}^0.
\end{dmath}

Besides the constraint on the rotation parameters, one must keep in mind that, for the point to be on the reflection curve, we have to take into account not only \eqref{eq:gama}, but also \eqref{eq:x_solved}. In order to account for \eqref{eq:x_solved}, another constraint is considered
\begin{equation}
  g_2(c\theta,s\theta,t_x,t_y) = \left\Vert x - \frac{c_{18}^2[c\theta,s\theta,t_x,t_y]}{c_{17}^1[c\theta,s\theta,t_x,t_y]} \right\Vert^2 = 0.
\end{equation}

Then, the absolute pose problem for NCCS, using 3D straight lines, is formulated as an optimization problem by taking the absolute value of the sum of the function $gama_r(c\theta,s\theta,t_x,t_y)$, for all matchings between 3D straight lines and the respective image pixels. The formal formulation is given by
\begin{equation}
  \begin{aligned}
    & \underset{c\theta,s\theta,t_x,t_y}{\text{min}}
    & & \frac{1}{NM}\sum_{i=1}^{N}\sum_{j=1}^{M} \left|\gamma_r(c\theta,s\theta,t_x,t_y)\right| \\
    & \text{s.t.} & & c_1(c\theta,s\theta) = 1 \\
    & & & c_2(c\theta,s\theta,t_x,t_y) = 0.
    \label{eq:final_problem}
  \end{aligned}
\end{equation}
To conclude, the rotation and translation is given by \eqref{eq:rt_para}, that satisfy \eqref{eq:final_problem}. 

\begin{figure}
  \centering
  \includegraphics[width=1\columnwidth]{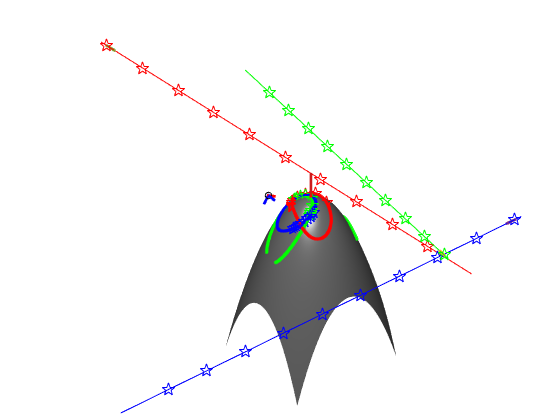}
  \caption{Straight lines projection onto the mirror of a non-central catadioptric camera. Filled lines in the mirror represent the reflection curve of each line. The star shape points on the straight lines represent the select points of each line to project using the method in Agrawal et. al.~\cite{article:agrawal:2011}, the points resulting from this method are plotted on the mirror's surface. The small axis represents the perspective camera COP.}
  \label{fig:proj_line}
\end{figure}


Most of the times, the pose is given by the rotation and translation that transform the points from the camera to the world coordinate systems. In order to obtain this transformation, one just needs to apply the inverse rigid transformation
\begin{equation}
  \mathbf{p}^{(\pazocal{W})} = \mathbf{R}^T\mathbf{p}^{(\pazocal{C})} - \mathbf{R}^T\mathbf{t},
\end{equation}
where $\mathbf{p}^{(\pazocal{W})}$ and  $\mathbf{p}^{(\pazocal{C})}$ represent a point in the world and camera reference frames respectively.

\section{Experimental Results}
\label{sec:results}

The proposed methods were evaluate by performing test with synthetic and real data. We used the Synthetic data to evaluate the performance of the pose estimation method in the presence of noise, both in the image of the lines points and in the 3D position of the lines. The synthetic data tests were performed for the 3 different types of mirrors, as define in \eqref{eq:mirror}. The parameters and the position of the COP, used in these experiments, are shown in Table~\ref{tab:para}. In addition (also with synthetic data), its performance for different number of lines is evaluated. The real data tests show an application of the pose estimation method to localize a mobile robot.

\begin{table}[]
\centering
\caption{Mirror Parameters and COP position for each mirror.}
\label{tab:para}
\begin{tabular}{l|r|r|r|}
\cline{2-4}
 & \multicolumn{3}{c|}{Mirror Type} \\ \hline
\multicolumn{1}{|c|}{Parameter} & \multicolumn{1}{c|}{Hyperbolic} & \multicolumn{1}{c|}{Parabolic} & \multicolumn{1}{c|}{Spheric} \\ \hline
\multicolumn{1}{|l|}{A} & -1.2 & 0 & 1 \\ \hline
\multicolumn{1}{|l|}{B} & 3.4 & 20.4 & 0 \\ \hline
\multicolumn{1}{|l|}{C} & -33.2 & 53.2 & 900 \\ \hline
\multicolumn{1}{|l|}{COP (x,y,z)} & (0, 25, 25) & (0, 30, 20) & (0, -15, 55) \\ \hline
\end{tabular}
\end{table}

\subsection{Using Synthetic Data}
\label{sec:syn_results}

\begin{figure*}
  \centering
  \subfloat[Box plot of the absolute rotation error in degrees for different levels of noisy pixels and three diferent mirrors.]{\includegraphics[width=0.48\textwidth]{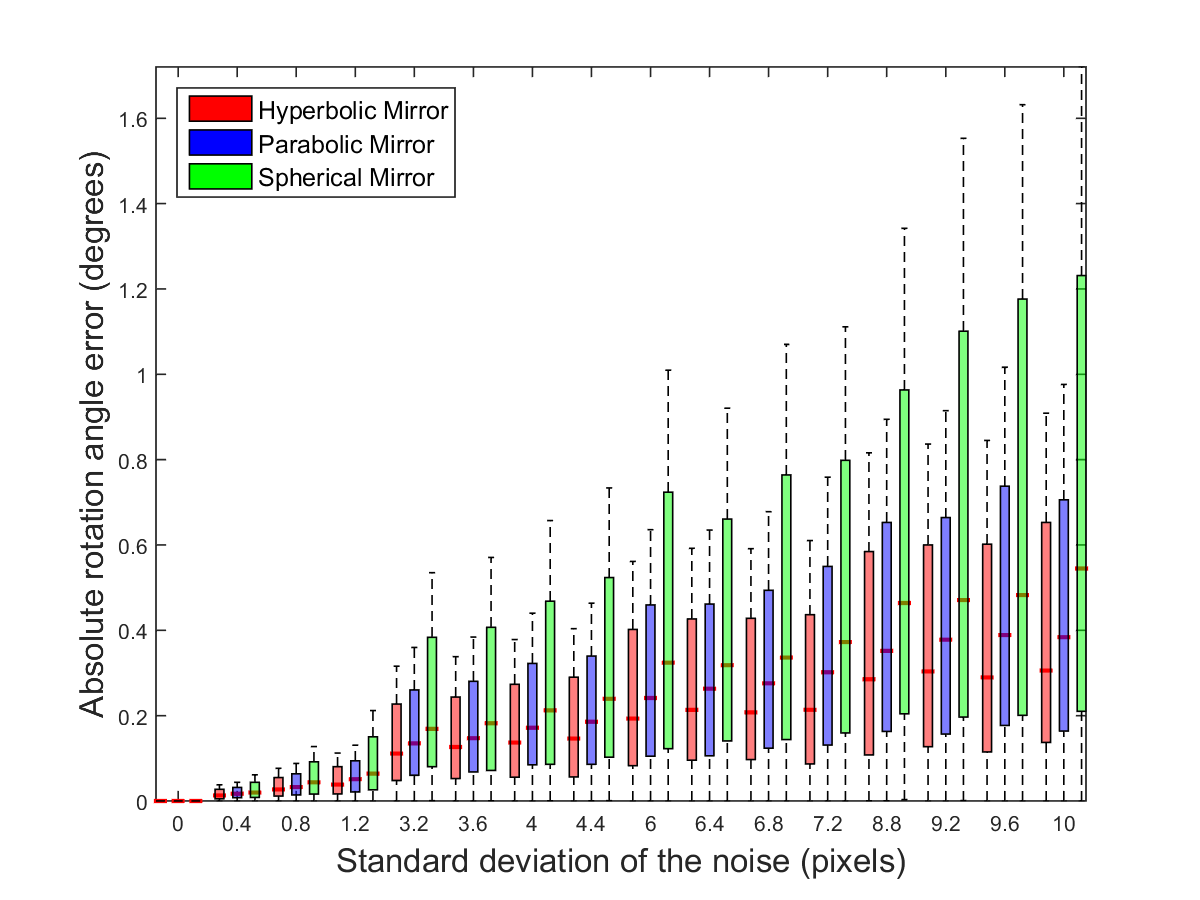}\label{fig:error_pixel:teta}} \hfill
  \subfloat[Box plot of the norm of the translation error in degrees for different levels of noisy pixels and three diferent mirrors.]{\includegraphics[width=0.48\textwidth]{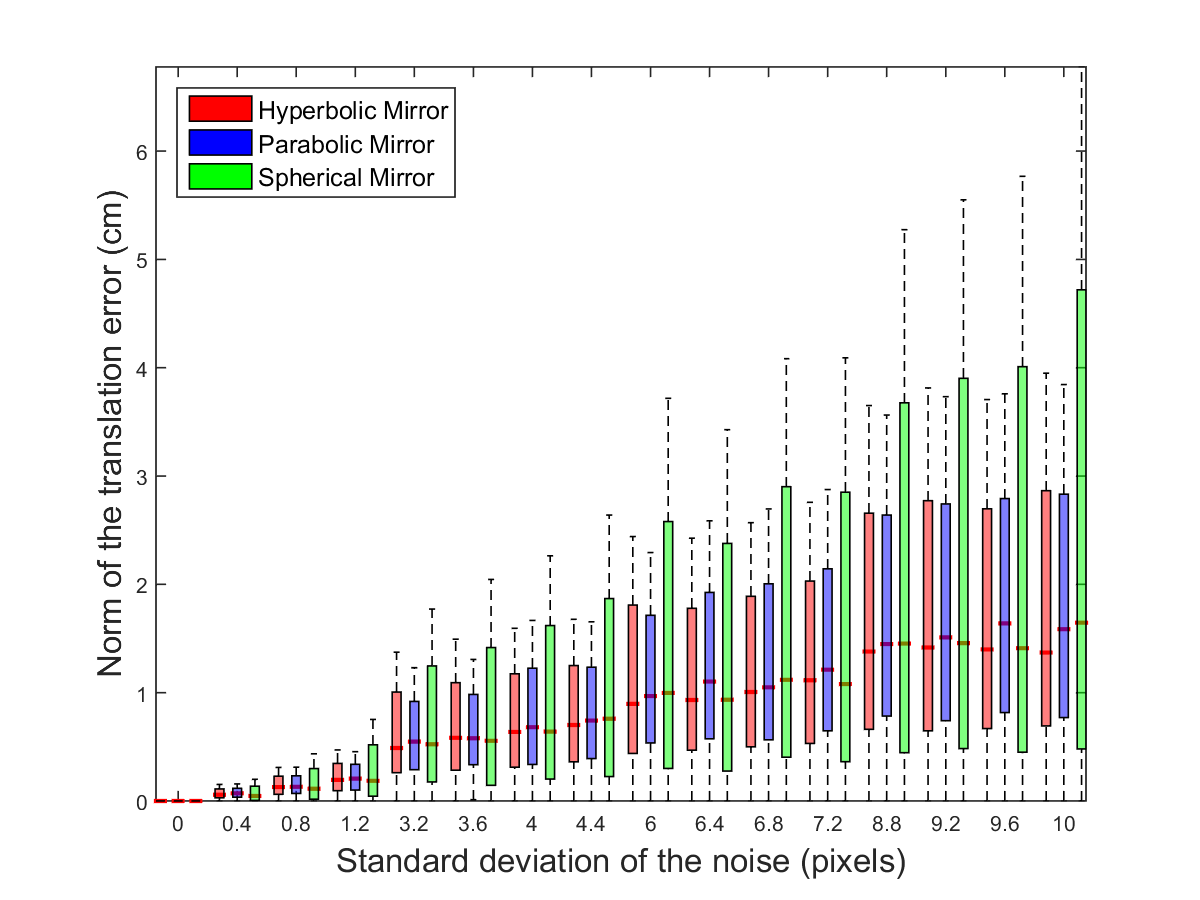}\label{fig:error_pixel:trans}} \hfill
  \caption{Method performance under noisy image pixels. For this experiment the data-set was generated for $M = 5$ and $N = 20$. The noise was introduced by adding samples from a normal distribution with zero mean and increasing standard deviation (x-axis) of the plots. The red lines in the box plots represent the median of the errors.}.
  \label{fig:error_pixel}
\end{figure*}

Before testing the pose estimation method, we validate the straight line projection equation, derived in Section~\ref{sec:line_proj}. In order to do that we defined a small set of lines, the position of the perspective camera (COP), the mirror parameters, and applied \eqref{eq:gama} and \eqref{eq:x_solved}. Afterwards, we selected a set of points of each line, applied the method proposed in Agrawal et. al.~\cite{article:agrawal:2011} and verified if the resulting points, where coincident with the previous computed reflection curves. In order to illustrate this results we plotted the lines, mirror, COP, and points using {\sc MATLAB}. The results can be seen in Fig.~\ref{fig:proj_line}.

Regarding synthetic data-sets, three tests were performed on the pose estimation method, to assess its performance. The first and second focused on evaluating the effects of noisy data in the final solution. The third consisted in evaluating the performance for different number of lines. Two different types of noisy data were considered, noise added in the line image pixels (first test), and noise added on the coordinates of the lines in the world reference frame (second and third tests). The data-sets were generated in {\sc Matlab} and the pose estimation algorithm was implemented using its optimization toolbox (code will be available on the authors page).

The procedure for generating the data-sets was as follows: a set of $N$ 3D straight lines were randomly generated, $\mathbf{p}_i(\lambda)^{(\pazocal{W})}$. Those lines were obtained by taking a set $N$ arbitrary points, $\mathbf{q}_i^{(\pazocal{W})}$, and directions, $\mathbf{d}_i^{(\pazocal{W})}$, with unit length, which define lines known to have a solution for the projection scheme described in Sec.~\ref{sec:line_proj}. To each point a random 3D rigid transformation, (defined by a random rotation matrix $\mathbf{R}_1$ and a random translation vector $\mathbf{t}_1$) is applied, to each direction a 3D random rotation ($\mathbf{R}_2$) is applied. Keep in mind that these rotation matrices and translation vector are independently generated (randomly) for each point and direction. Each line will then be defined by
\begin{equation}
  \mathbf{p}_i(\lambda)^{(\pazocal{W})} = \lambda\mathbf{R}_2\mathbf{d}_i^{(\pazocal{W})} + \mathbf{R}_1\mathbf{q}_i^{(\pazocal{W})} + \mathbf{t}_1.
  \label{eq:rand_lines}
\end{equation}

The lines defined by \eqref{eq:rand_lines} are then transformed by the random ground truth rotation and translation parameters \eqref{eq:rt_para}. From the resulting lines, a set of $M$ points per line are selected and projected to the mirror using the method in \cite{article:agrawal:2011}, yielding the points, $\mathbf{m}_{ij}^{(\pazocal{C})}$, which represent the projection of the $j^{th}$ point of the $i^{th}$ line.

The goal of the first test was to evaluate the method performance in the presence of noisy data. In this test, noise was added to pixels of the images of each line point. Given that, the camera's intrinsic parameters were known, the process of adding noise to the pixels was straight forward. The first step was projecting the set of points $\mathbf{m}_{ij}^{(\pazocal{C})}$ to the image plane. 
 Then to the resulting pixels were added samples from a normal distribution with zero mean and standard deviation ranging from $0$ to $10$. Finally, the pixels were re-projected onto the mirror by intersecting the resulting directions (inverse projection of the camera's pixels) with the known mirror equation \eqref{eq:mirror}. For each value of the standard deviation, $1000$ trials were performed. Henceforward consider a trial to be the execution of the method for a data-set generated as described previously. Results for three different types of mirrors are presented in Fig.~\ref{fig:error_pixel}\subref{fig:error_pixel:teta} and Fig.~\ref{fig:error_pixel}\subref{fig:error_pixel:trans}.

\begin{figure*}
  \centering
  \subfloat[Box plot of the absolute rotation error in degrees for different levels of noisy world points.]{\includegraphics[width=0.48\textwidth]{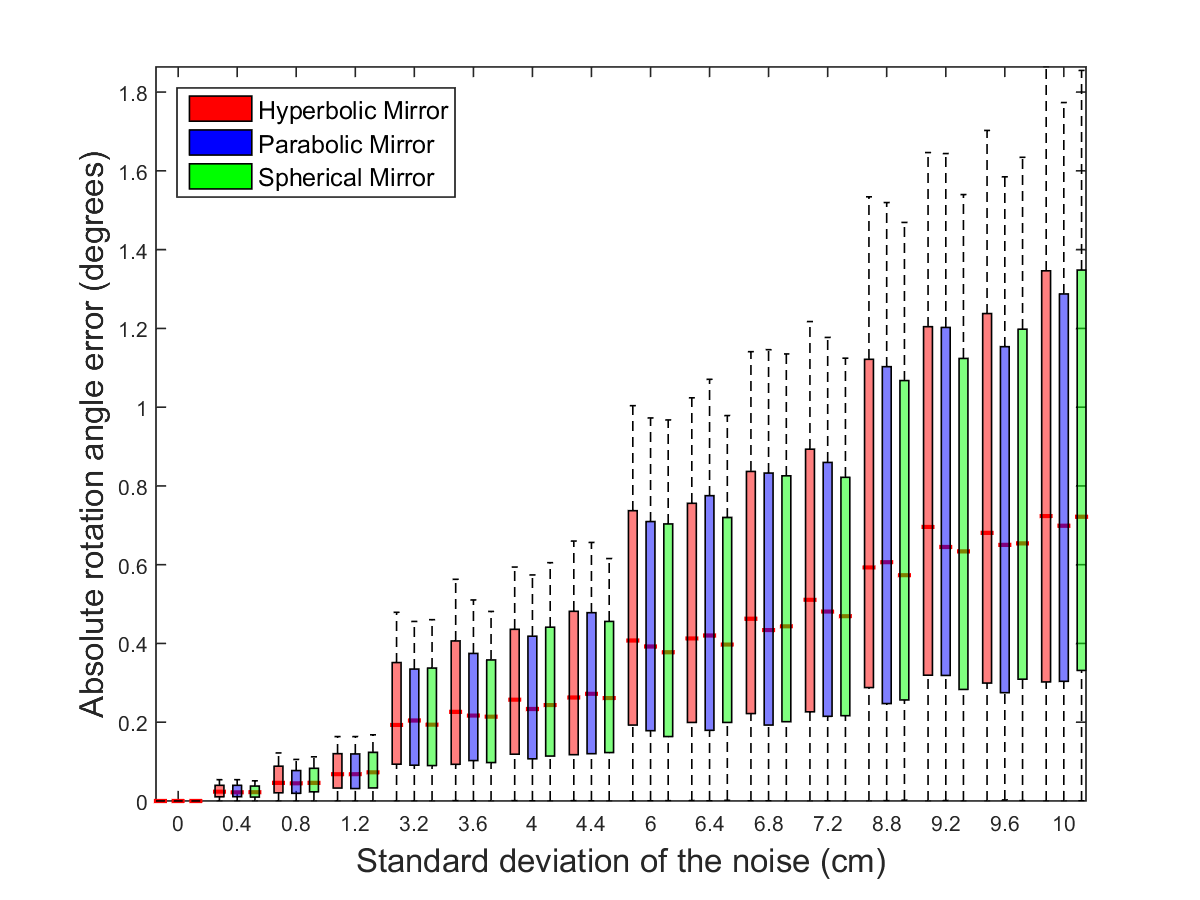}\label{fig:error_world:teta}} \hfill
  \subfloat[Box plot of the absolute rotation error in degrees for different levels of noisy world points.]{\includegraphics[width=0.48\textwidth]{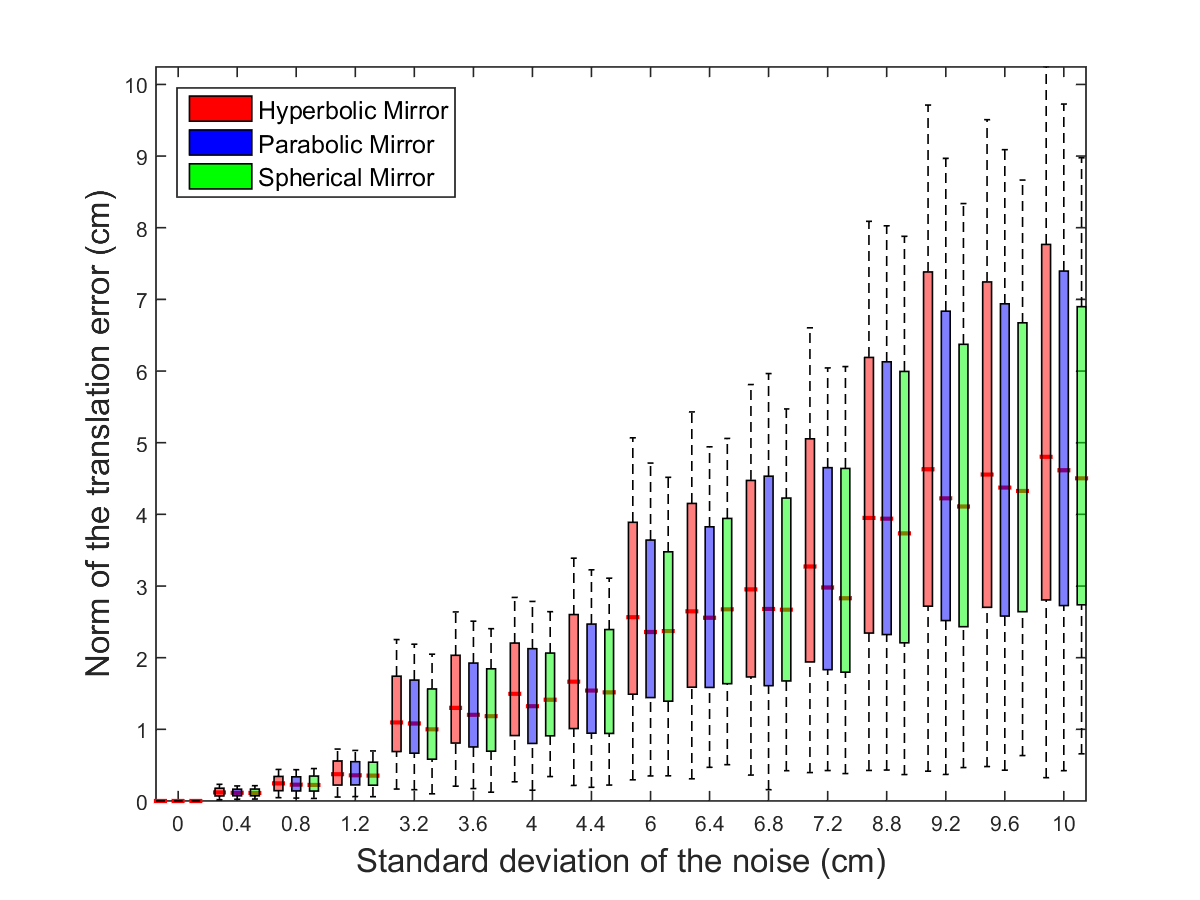}\label{fig:error_world:trans}} \hfill
  \caption{Method performance under noisy world points. For this experiment the data-set was generated for $M = 10$ and $N = 20$. The noise was introduced by adding samples from a normal distribution with zero mean and increasing standard deviation (x-axis) of the plots. The red lines in the box plots represent the median of the errors.}
  \label{fig:error_world}
\end{figure*}

The second test consisted on adding noise to the straight lines points before projecting them onto the mirror. The noise is introduced by adding samples of a normal distribution with zero mean and standard deviation ranging from $0$ to $10$. Afterwards they are projected onto the mirror and the resulting points are the ones used to estimate the pose. The results for the absolute rotation angle and the norm of the translation error are presented in Fig.~\ref{fig:error_world}\subref{fig:error_world:teta} and Fig.~\ref{fig:error_world}\subref{fig:error_world:trans}.

\begin{figure*}
  \centering
  \subfloat[Box plot of the absolute rotation error in degrees for different number of lines.]{\includegraphics[width=0.4\textwidth]{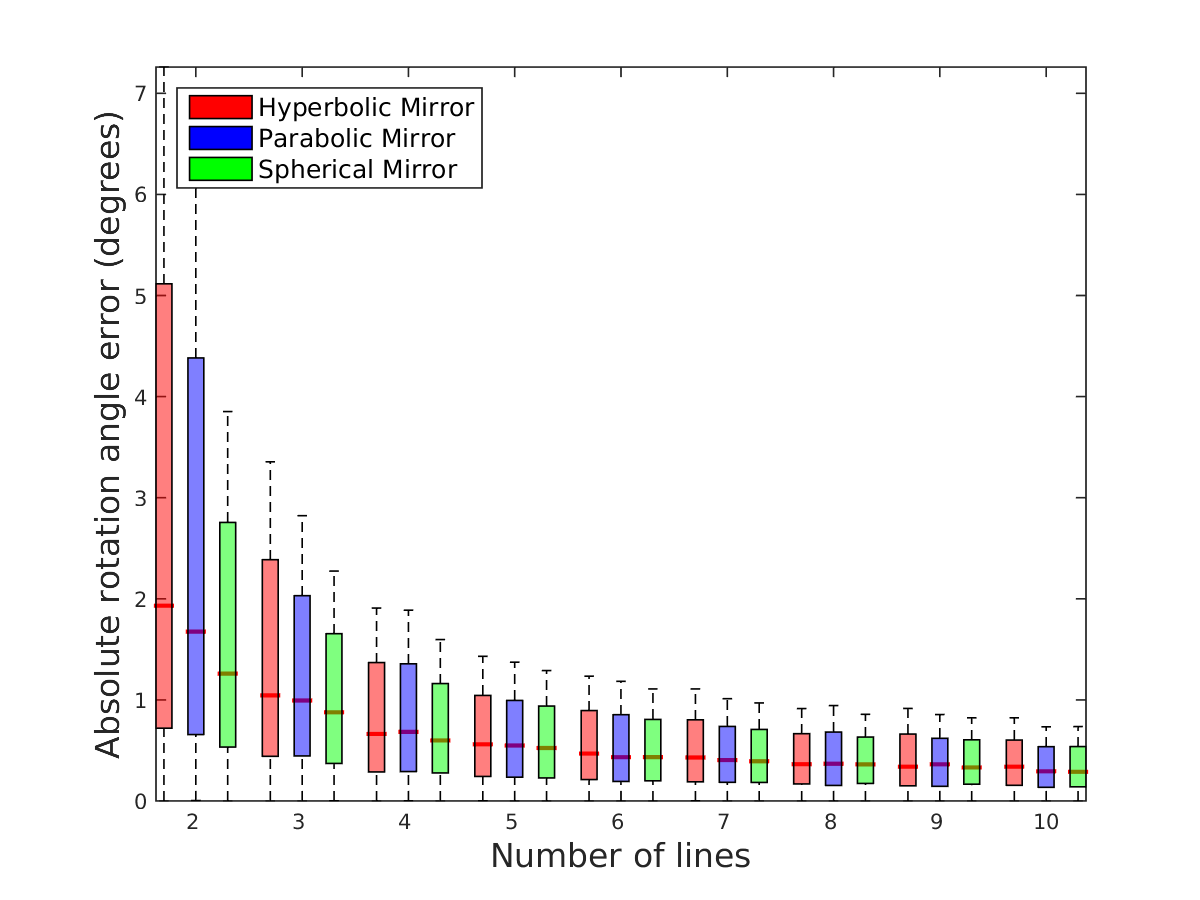}\label{fig:error_nlines:teta}} \hfill
  \subfloat[Box plot of the absolute rotation error in degrees for different number of lines.]{\includegraphics[width=0.4\textwidth]{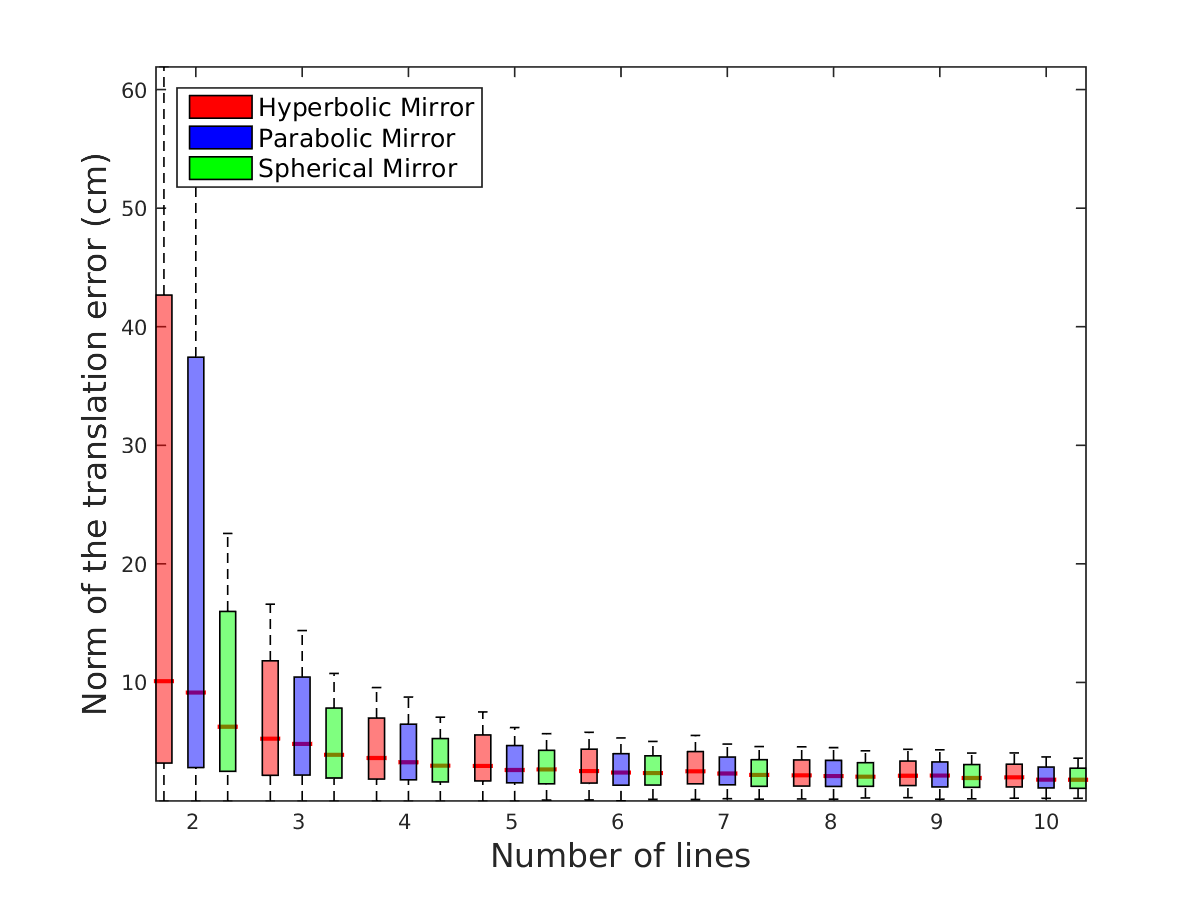}\label{fig:error_nlines:trans}} \hfill
  \caption{Method performance under noisy world points. For this experiment the data-set was generated for different number of lines $M$ and $N = 20$. The noise was introduced by adding samples from a normal distribution with zero mean and standard deviation of $5cm$ (x-axis) of the plots. The red lines in the box plots represent the median of the errors.}
  \label{fig:error_nlines}
\end{figure*}

Finally the third test was similar to the second, with the difference that the noise standard deviation was fixed at $5cm$, and what varies throughout the trials is the number of lines $M$ used by the method. The results for the absolute rotation angle and the norm of the translation error are presented in Fig.~\ref{fig:error_nlines}\subref{fig:error_nlines:teta} and Fig.~\ref{fig:error_nlines}\subref{fig:error_nlines:trans}.

\subsection{Using Real Data}
\label{sec:real_results}

The application considered for the real data experiments was visual navigation. For that purpose we mounted a non-central catadioptric camera, composed by a perspective camera and a spherical mirror, on top of a Pioneer-3DX robotic platform. The NCCS was calibrated with the method proposed at Perdigoto and Araujo \cite{article:perdigoto:2013}.

\begin{figure*}
  \centering
  \subfloat[Non-central catadioptric camera image, with the line detected marked with different colors.]{\includegraphics[width=0.48\textwidth]{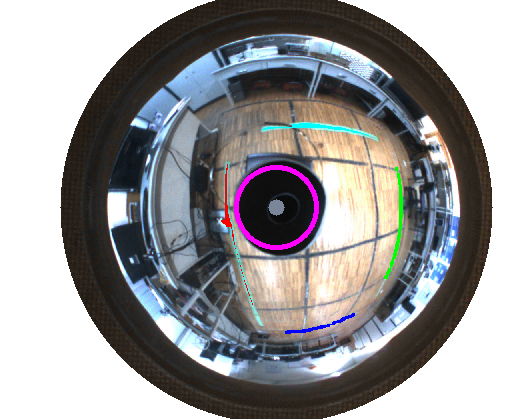}\label{fig:cam:initial}} \hfill
  \subfloat[Non-central catadioptric camera image from another view, with the line detected marked with different colors.]{\includegraphics[width=0.48\textwidth]{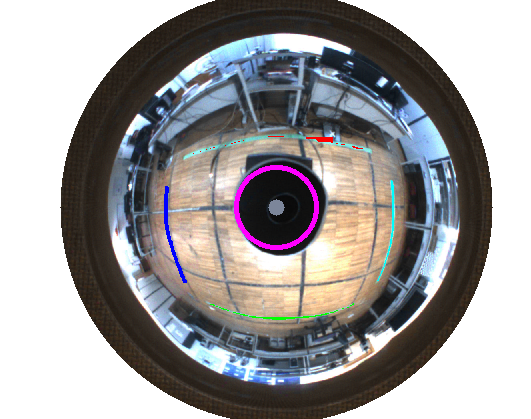}\label{fig:cam:rotate}} \hfill
  \caption{Two images from the NCCS mounted on top of the mobile platform throughout one real data experiment. The lines marked in the images were the detected with the method described in Section~\ref{sec:real_results}.}
  \label{fig:cam}
\end{figure*}

The 3D world lines considered were four green lines on the ground. In order to generate a data-set, we need to associate pixels in the image to the lines, that association is performed in four steps. First we apply a color threshold (in this case green) to the image and then morphological operators to remove noise; then find blobs by close contour extraction. The first image is used as reference to associate manually the blobs with the 3D lines. For consequent images the process is automatic. Finally we take $75$ pixels of each line image from the associated blobs. Images from the camera in the setup and the lines detected with the method previously described are presented in Fig.~\ref{fig:cam}\subref{fig:cam:initial} and Fig.~\ref{fig:cam}\subref{fig:cam:rotate}
 
Given that the method requires the mirror points as well, we use the scheme previously discussed. The projection line for each pixel is computed and intersected with the mirror \eqref{eq:mirror}. All the image processing steps were implemented in C++, using OpenCV. 

The optimization step was implemented using the {\sc MATLAB}'s Optimization Toolbox. The communication between the camera, the line-pixel association, the optimization software and the robot was handle resorting to the Robot Operative System (ROS) topic API. The results of this test were recorded in a video, consisting of the robot's pose throughout the execution of a trajectory. This video will be sent as supplementary material.

\section{Conclusions}
\label{sec:conclusions}

\subsection{Analysis of the Experimental Results}

This section presents the analysis of the experimental results of the pose estimation method presented in Sections~\ref{sec:syn_results} and \ref{sec:real_results}.

Starting by the synthetic data experiments. Three different tests were performed. The first consisted on assessing the method performance in presence of noisy images. As we can see from Fig.~\ref{fig:error_pixel}\subref{fig:error_pixel:teta} the method proved to be robust to noisy images, in the rotation estimation, with the median of the absolute rotation angle error never going above $0.6$ degrees for every standard deviation value and for all mirrors. The translation error was measured by the norm of the difference between the ground truth and the estimated translations in both directions ($x$ and $y$). Again the results for the translation prove the robustness of the method with the maximum median of the error being less than $2$cm, for a noise value of $10$ pixels.

Then the performance under noise in the 3D lines was evaluated. In Fig.~\ref{fig:error_world}\subref{fig:error_world:teta} we present the absolute rotation angle error evolution with the increasing of the noise standard deviation. It can be seen that the method still presented a good performance, however it presented a slightly higher error than for the pixel noise. As far as the norm of translation error, Fig.~\ref{fig:error_world}\subref{fig:error_world:teta}, the same behavior was seen, the median of the error is small, but is slightly higher than the previous experiment.

The last test with synthetic data evaluated how the number of 3D lines considered in the method affect its performance. As expected and shown in Fig.\ref{fig:error_nlines}\subref{fig:error_nlines:teta} and in Fig.Fig.\ref{fig:error_nlines}\subref{fig:error_nlines:trans}, both the rotation and translation decrease considerably when the number of lines increases.

The real data experiments, as seen in the video (sent as supplementary material) the robot exhibited a good performance. Keep in mind that the only sensor used throughout this experiments was the NCCS.

Finally a brief comment on the convergence of the method. The method converges even for initial values distant from the optimal value, however the computation time increases as the initial value e set further away from the optimal. This is expected, since the optimizer will need to compute more iterations to reach the optimal value. This was seen especially in the synthetic data tests. In the real data experiments this problem did not had a high influence in the computation time, because the initial value in each time step was set to the previous estimated pose.

\subsection{Closure}

In this article we purpose a novel method for planar pose estimation of mobile robots. This method is specific to non-central catadioptric cameras and is based on the projection of 3D straight lines onto the mirror of those devices. This projection is given by a $10^{th}$ polynomial equation, whose derivation we also presented in this paper. This equation is, then, rewritten as a function of the rigid transformation parameters and used to formulate an optimization problem for the pose estimation.

The pose estimation method was validated with synthetic and real data. The former proved that the method is robust to the presence of noise both in the 3D lines and in the image pixels. Besides, they showed that the performance of the method increases significantly with the number of lines considered. Finally, we showed the method performance in a visual navigation context with a real robot.  

\bibliographystyle{IEEEtran}
\bibliography{root}

\end{document}